\documentclass[letterpaper, 10 pt, conference]{ieeeconf}  

\IEEEoverridecommandlockouts                              

\usepackage{graphics} 
\usepackage{epsfig} 
\usepackage{mathptmx} 
\usepackage{times} 
\usepackage{amsmath} 
\usepackage{amssymb}  
\usepackage{url}
\usepackage{multirow}
\usepackage[table,xcdraw]{xcolor}

\title{\LARGE \bf
CognitiveOS: Large Multimodal Model based System to Endow Any Type of Robot with Generative AI
}

\author{
  Artem Lykov$^{1}$,
  Mikhail Konenkov$^{1}$,
  Koffivi Fidèle Gbagbe$^{1}$,
  Mikhail Litvinov$^{1}$,\\
  Denis Davletshin$^{1}$,
  Aleksey Fedoseev$^{1}$,
  Miguel Altamirano Cabrera$^{1}$,
  Robinroy Peter$^{1}$,\\
  and Dzmitry Tsetserukou$^{1}$\\
  \thanks{$^{1}$ISR Lab, Skolkovo Institute of Science and Technology, Moscow, Russia {\tt\small \{Artem.Lykov, Mikhail.Konenkov, Koffivi.Gbagbe, Mikhail.Litvinov2, Denis.Davletshin, Aleksey.Fedoseev, M.Altamirano, Robinroy.Peter, D.Tsetserukou\}@skoltech.ru}}
}

\begin{document}

\maketitle
\thispagestyle{empty}
\pagestyle{empty}

\begin{abstract}

This paper introduces CognitiveOS, the first operating system designed for cognitive robots capable of functioning across diverse robotic platforms. CognitiveOS is structured as a multi-agent system comprising modules built upon a transformer architecture, facilitating communication through an internal monologue format. These modules collectively empower the robot to tackle intricate real-world tasks. The paper delineates the operational principles of the system along with descriptions of its nine distinct modules. The modular design endows the system with distinctive advantages over traditional end-to-end methodologies, notably in terms of adaptability and scalability. The system's modules are configurable, modifiable, or deactivatable depending on the task requirements, while new modules can be seamlessly integrated. This system serves as a foundational resource for researchers and developers in the cognitive robotics domain, alleviating the burden of constructing a cognitive robot system from scratch. Experimental findings demonstrate the system's advanced task comprehension and adaptability across varied tasks, robotic platforms, and module configurations, underscoring its potential for real-world applications. Moreover, in the category of Reasoning it outperformed
CognitiveDog (by 15\%) and RT2 (by 31\%), achieving the highest to date rate of 77\%. We provide a code repository and dataset for the replication of CognitiveOS: \textbf{link will be provided in camera-ready submission.}

\end{abstract}


\section{Introduction}

The growing popularity of transformer-based Large Language Models (LLMs) such as OpenAI's ChatGPT \cite{lib:openai2022introducing} and subsequent open-source models like LLaMa \cite{lib:touvron2023llama}, Falcon \cite{almazrouei2023falcon}, and Mistral \cite{jiang2023mistral}, has significantly contributed to the advancement of artificial intelligence across various technological domains. In cognitive robotics, the scientific community recognized the high generalization capability of these large models as a key to developing a robot that could perform new tasks based on generalized knowledge derived from familiar actions expressed in natural language. However, efforts to apply LLMs in robotics face challenges, particularly in understanding and processing the external world.

\begin{figure}[thpb]
    \centering
    \vspace{0.2cm}
    \includegraphics[scale=0.175]{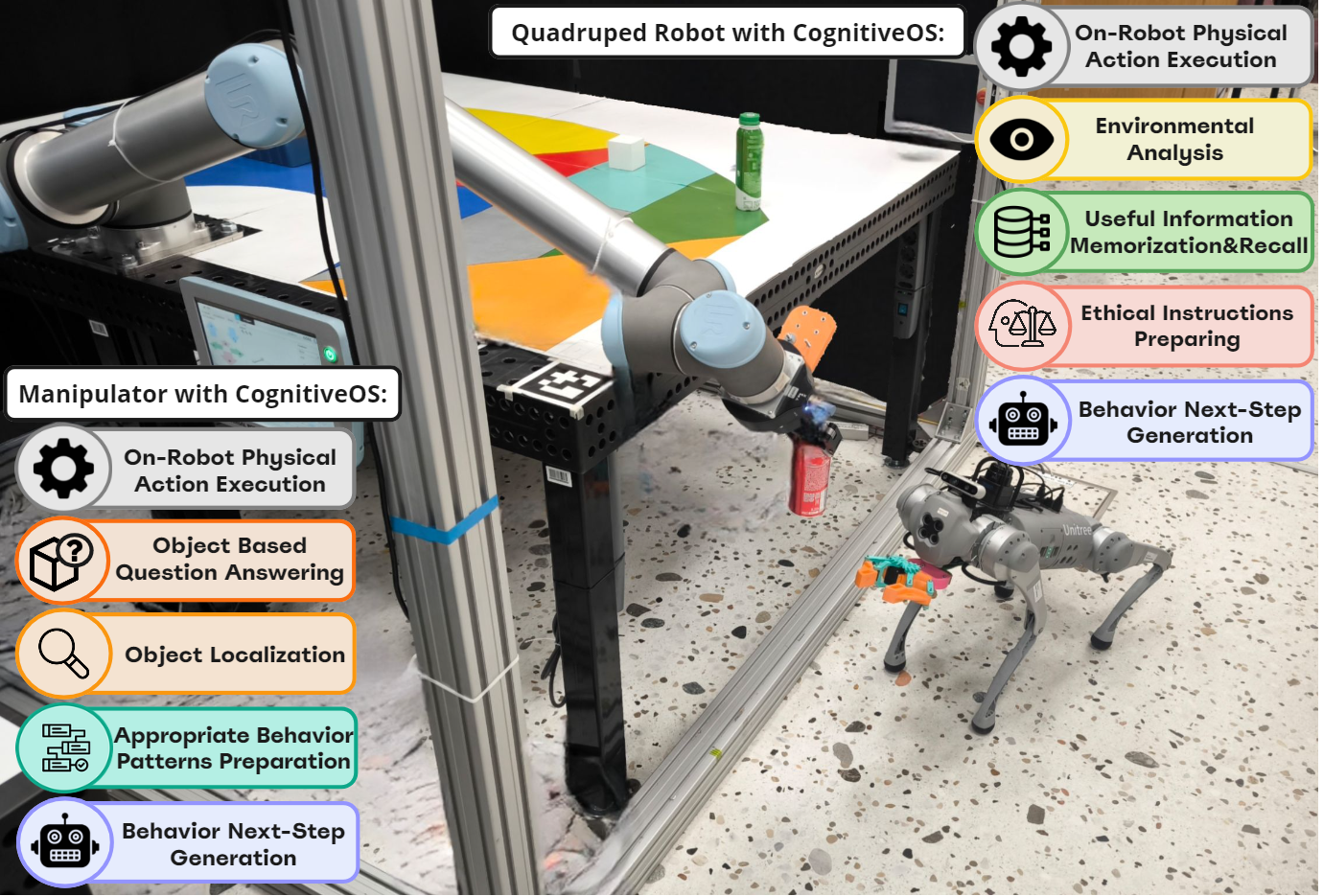}
    \caption{Application of CognitiveOS to Various Robotics Platforms.}
    \label{figurelabel}
    \vspace{-0.6cm}
\end{figure}

Previous attempts to convey the model's understanding of the world through text-only approaches \cite{ahn2022can}, \cite{lib:singh2023progprompt}, \cite{hao2023reasoning} struggled with ambiguities and the assumption of static objects unless interacted with. The introduction of multi-modal transformer-based models such as GPT-4 \cite{openai2023gpt4} and Gemini \cite{Pichai2023}, capable of processing images, opened up new possibilities for robotics \cite{driess2023palm}, allowing robots to comprehend their environment and enhancing their "Embodied Experience" \cite{mu2023embodiedgpt}.

Cognitive robots have been developed on various platforms, ranging from mobile manipulators \cite{driess2023palm}, \cite{brohan2023rt} to bio-inspired humanoid robots \cite{Tesla_bot}, \cite{DIGIT} and Quadruped robots \cite{lib:bostondynamics2023robotschat}, \cite{lykov2024cognitivedog}. In the latter, cognitive abilities were developed using an "Inner Monologue" approach \cite{lib:huang2022inner}, with improvements inspired by the "Autogen" concept \cite{wu2023autogen}. The cognition of the robot is facilitated through internal communication between agent models, leveraging their strengths to provide different cognitive capabilities to the system. 

In this paper, we present a novel approach to cognitive robotics, focusing on the development of a scalable modular cognitive architecture (SMCA), each module of which can be modified, disabled, or the system can be extended with additional modules. This gives the possibility to add new cognitive abilities by adding new modules to the architecture, as well as to run SMCA on different robotic platforms as it is a replacement for the On-Robot Physical Action Execution module. All of these modifications can be made without the need for a major system change. This modular architecture creates a scalable tool for creating cognitive robots on any platform, ensuring flexibility and adaptability in robotic cognition. Additionally, our approach offers the potential to provide several different robots with the necessary cognitive abilities through a single system running on a server.

\vspace{-0.1cm}
\section{Related Works}

The field of robotic AI has witnessed notable advancements that closely align with our research. Key contributions include:

\textbf{Human Task Understanding in Manipulator Robots:}
Earlier works showcased robots capable of comprehending and executing human tasks PaLM-E \cite{driess2023palm}, RT1 \cite{brohan2022rt}, RT2 \cite{brohan2023rt}. These models, akin to LLM outputs, improved performance through an inner monologue approach \cite{lib:huang2022inner}, refining actions based on task feedback. Microsoft Autogen project \cite{wu2023autogen} represents a shift from single-model inner monologues to multi-model conversations. 

\textbf{Multi-Model Conversations in Cognitive Robotics}:
In cognitive robotics, the interaction of multiple models in an internal monologue has been used in the work of CognitiveDog \cite{lykov2024cognitivedog}. This system allows a robot to communicate with humans and physically interact with the environment by manipulating objects. The system is implemented on a quadruped robot and demonstrates autonomous decision-making capabilities by autonomously determining the most appropriate actions and interactions with various objects to accomplish user-defined tasks.

\textbf{Mistral 7B Model Integration}:
The Mistral 7B model excels in MMLU, commonsense reasoning, and world knowledge, outperforming LLaMa2 13B in most evaluations. Its superiority makes it an ideal candidate for generating robotic actions, considering environmental context and historical data.

\textbf{Qwen-VL and Environmental Analysis}:
Qwen-VL \cite{bai2023qwen} stands out in the field of Visual Language Models (VLM), demonstrating a remarkable performance in image captioning, question answering, visual localization, and flexible interaction tasks. It accepts text and images as input. Qwen-VL also excels in recognizing text and symbols and grounding tasks, which is important for our design. This model's strength lies in the usage of a position-aware vision-language adapter and a 3-stage training pipeline. QwenVL has a total of 9.6B parameters, which is the sum of the parameters in a visual encoder (1.9 billion), a vision-language adapter (0.08 billion), and a LLM (Qwen, 7.7 billion).

\begin{figure}[thpb]
    \centering
    \vspace{0.2cm}
    \includegraphics[scale=0.20]{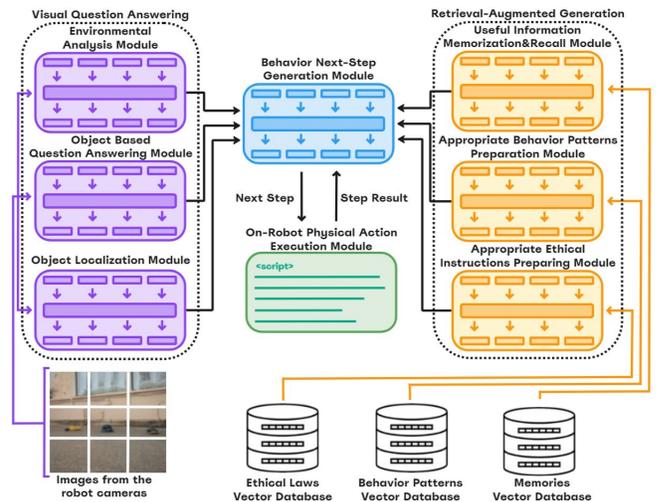}
    \caption{System Architecture.}
    \label{figurearc}
    \vspace{-0.5cm}
\end{figure}

\section{Cognitive Operating System Architecture}

This section provides a comprehensive System Overview of our approach, with the system architecture shown in Fig.  \ref{figurearc}. A key feature distinguishing our approach from other developments in the field of cognitive robotics is its modularity and scalability. This characteristic enables the modification, removal, and addition of modules to the system, rendering it applicable to a wide range of robotic platforms. This achievement is attributed to the standardization of the system's inter-module interaction in the form of an internal dialogue. To achieve the user-defined goal incrementally, the robot's behavior is constructed step by step. Each subsequent step is formulated considering all previous steps, their outcomes, and the information transmitted by all modules, the expert domains of which may be relevant in this context. All modules except “On-Robot Physical Action Execution Module" are based on large transformer-based models and Python shells. The modules are functionally separated from the system according to the principle of involvement in solving one or another cognitive task. Some functional modules can be combined in one large model if there is a model capable of performing several tasks for more economical use of computational resources. Below is a list of modules implemented during system development along with their descriptions.

\subsection{Behavior Next-Step Generation Module}

\begin{figure}[thpb]
    \centering
    \vspace{0.2cm}
    \includegraphics[scale=0.40]{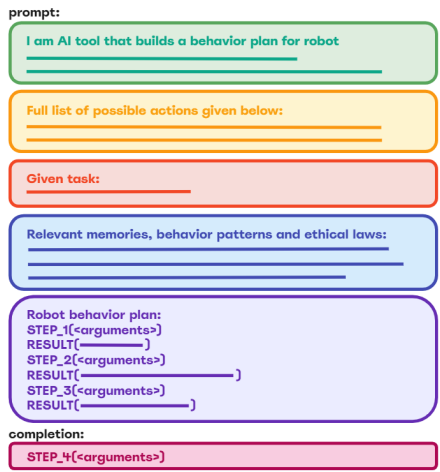}
    \vspace{0.2cm}
    \caption{Prompt structure of the Behavior Generation Module.}
    \label{figureprompt}
    \vspace{-0.35cm}
\end{figure}

This module serves as the cornerstone of the robot's cognitive system and is the only one that must be present in the system in some form. It is responsible for the step-by-step generation of the robot's behavior based on all received external and internal information. The module is implemented based on the LLM Mistral 7B, fine-tuned on a custom dataset of examples of behaviors from various robots. The objective of this module is to generate the next step of the robot in the current situation based on all available information. Some information is requested from other modules immediately after receiving the task from the user, while other information is requested from modules during the task execution using corresponding behavior steps. An example of how a prompt for the LLM in the Behavior Generation Module might appear is illustrated in Fig. \ref{figureprompt}. If deciding the next step requires additional thought, the model can break it down and solve it step by step iteratively generating step \textbf{“THOUGHT"}. If the data indicates task completion, the model produces \textbf{“FINISH"} as the next step.

\subsection{On-Robot Physical Action Execution Module}

This module is equally critical as it governs the execution of physical actions on the robot. It facilitates physical actions on the robot such as object manipulation, movement, and other actions within the scope of actions available on the robotic platform. The list of physical actions implemented on various robots during the system development includes: \textbf{“SAY”}, \textbf{“LISTEN”}, \textbf{“GO\_TO”}, \textbf{“SIT”}, \textbf{“UP”}, \textbf{“TURN”}, \textbf{“FOLLOW”}, \textbf{“TILT”}, \textbf{“DANCE”}, \textbf{“GO\_USER”}, \textbf{“TAKE”}, \textbf{“PUT\_IN”}, and \textbf{“GIVE\_TO\_USER”}.

Upon completing an action on the robot, the On-Robot Physical Action Execution Module returns its result to the Behavior Generation Module. Formalizing and separating the On-Robot Physical Action Execution into a distinct module has rendered it replaceable, along with the robotic platform on which the system operates. When replacing the platform, it is imperative to maintain the format of information exchange and provide the system with an updated list of available actions on the robot. The quality of task execution by the system in this scenario depends on the quality of implementation of these actions on the robot.

\subsection{Environmental Analysis Module}

The next module is built upon a VLM and serves to analyze the surrounding environment. Such a module can be effectively implemented based on open-source Visual Question Answering (VQA) models without fine-tuning, provided they give detailed responses to questions accurately. During the development of the module, promising results were achieved with models such as MiniGPT4-v2 \cite{chen2023minigptv2}, CogVLM \cite{hong2023cogagent}, and Qwen-VL \cite{bai2023qwen}. The system utilizes this module when a comprehensive analysis of the environment or its elements is needed, for instance, describing its surroundings or a picture in front of it. In such cases, the Behavior Generation Module generates a step \textbf{“DESCRIBE\_VIEW”} with a question as an argument. This question is passed to the Environmental Analysis Module along with the image from the robot's camera, and the response is returned as the result of this step.

\subsection{Object Based Question Answering Module}

The Object Based Question Answering Module is also based on a VLM, but its scope of application differs. As many robot tasks involve interaction with the surrounding environment and object manipulation, a separate tool is necessary for object recognition. This module is fine-tuned to provide concise answers to questions about objects. For instance, using the \textbf{“QUESTION\_VIEW”} step, one can inquire about the presence of objects of a certain type, class, or color in the environment, determine which object possesses certain qualities, or have a certain location description. We have fine-tuned the base VLM for this module not only to improve the quality of answers regarding objects but also to standardize the output format and avoid unnecessary information in the response.

\subsection{Object Localization Module}

Another module based on a VLM is fine-tuned for object detection in images captured by the robot's camera. The Object Localization Module enables the Behavior Generation Module to generate the \textbf{“SEARCH\_VIEW”} step. This step takes as input the name of the object or its brief description and returns a pointer to the found object and its bounding box on the image, which is then used for 3D localization. Typically, before searching for an object in the robot's behavior, all necessary guiding questions are asked using \textbf{“QUESTION\_VIEW”}, allowing for a concise and clear request for object search. Objects found in this manner can then be used for robot navigation or for physical interaction with them, depending on the capabilities of the robotic platform.

\subsection{Useful Information Memorization\&Recall Module}

This module utilizes Retrieval Augmented Generation technology \cite{lewis2021retrievalaugmented}, \cite{gao2024retrievalaugmented} and acts as the robot's long-term memory. It performs two key functions: memorizing and recalling useful information. To memorize information, the module retrieves information from the results of the robot's behavioral steps. The LLM selects information to be stored in a memory vector database. For recall, the module generates a short list of information useful for the task assigned to the robot based on the information stored in the vector database. This information is then included in the behavior generation module's prompts during the generation of behavior steps.

\subsection{Appropriate Behavior Patterns Preparation Module}

Another module employing Retrieval-Augmented Generation (RAG) technology has been developed for preparing behavior patterns useful in current tasks. Through working with the Behavior Generation Module, it was determined that One Shot Learning significantly enhances the quality of behavior plan generation for the robot. The prompt for generating the behavior plan contains several examples of behaviors for similar tasks, and the LLM constructs a plan based on their similarity. Suitable examples are selected from a pre-formed vectorized database of behavior patterns. Integration of this module into the system enhances its stability and enables the addition of new behaviors without the need to retrain the model. To do this, previously unknown behaviors need to be added to the behavior patterns vector database.

\begin{figure}[ht]
 \vspace{0.4cm}
  \includegraphics[width=0.49\textwidth]{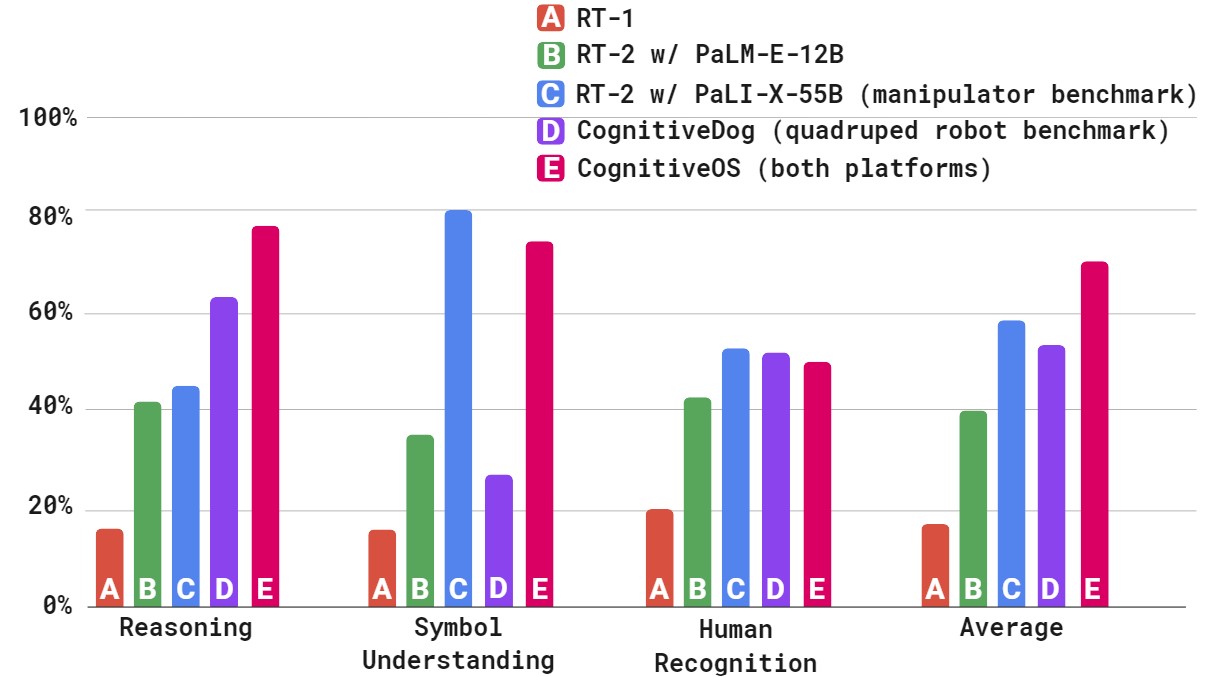}
  \caption{Performance comparison for three categories of emergent capabilities for different types of robotic platforms.}
  \label{fig:res1}
\end{figure}

\subsection{Appropriate Ethical Instructions Preparing Module}

Another module utilizing Retrieval-Augmented Generation (RAG) specializes in ethics. This time, a vectorized database stores a set of ethical rules that the robot must adhere to. The LLM generates recommendations for task execution based on these rules. These recommendations are then added to the prompt of the Behavior Generation Module and taken into account when selecting subsequent steps. The set of ethical rules can also be reviewed for each type and application area of robots.

\section{Evaluation}

During the development phase, the system was deployed and tested on various robotic platforms with ROS2 \cite{macenski2023survey}, including quadruped robots, quadruped robots with grippers, and static six-degree-of-freedom robotic manipulators. For the quadruped platform, we utilized the Unitree Go1 Edu robot \cite{unitreeleggedsdk} equipped with LIDAR and an RGB-D camera (Intel RealSense D435i). For the robotic manipulator, we employed the Universal Robot UR10, equipped with a 2-finger robotic gripper 2F-85 by Robotiq and a static RealSense 435i camera (1920 × 1080, 30 fps) for object localization within its workspace. Each of these robots had its On-Robot Physical Action Execution Module and a corresponding list of available actions determined by the physical capabilities of the platform. Some modules of the system were disabled and re-enabled during testing to evaluate the scalability of the system. Additionally, virtual robots with different parameters were employed during testing to increase the representativeness of the sample.

\subsection{System Performance Across Robots}

The aim of this experiment is to evaluate the system's ability to generalize knowledge and skills across different robotic platforms. We adopted the methodology proposed in RT2 and later in CognitiveDog. Tasks were categorized into Reasoning, Human Recognition, and Symbol Understanding categories. RT2 served as benchmark for the robot arm, while CognitiveDog provided benchmarks for the robot dog. CognitiveOS has been evaluated on both of these platforms. The experimental results are depicted in Fig. \ref{fig:res1}.

The experiment showed promising results. The system performance is platform-independent. This feature is provided by adapting the On-Robot Physical Action Execution module for each platform. On average, the result of CognitiveOS for both platforms outperforms benchmarks (for quadruped robot and manipulator by 15\% and 10\%, respectively). For Symbol Understanding the system came close to RT2 readings not getting to it only 8\% and surpassed CognitiveDog more than 3 times (74\% vs. 24\%).  In the Human Recognition category, CognitiveOS is almost on the same level as benchmarks. The main finding of the experiment is the excellent results of CognitiveOS in the Reasoning category. It is this parameter that is key to reflecting the intellectual abilities of a cognitive robot. In this category, it managed to outperform CognitiveDog and RT2 by 15\% and 31\%, respectively, and to achieve the previously impossible 77\%.

\begin{figure}[ht]
 \centering
  \includegraphics[width=0.45\textwidth]{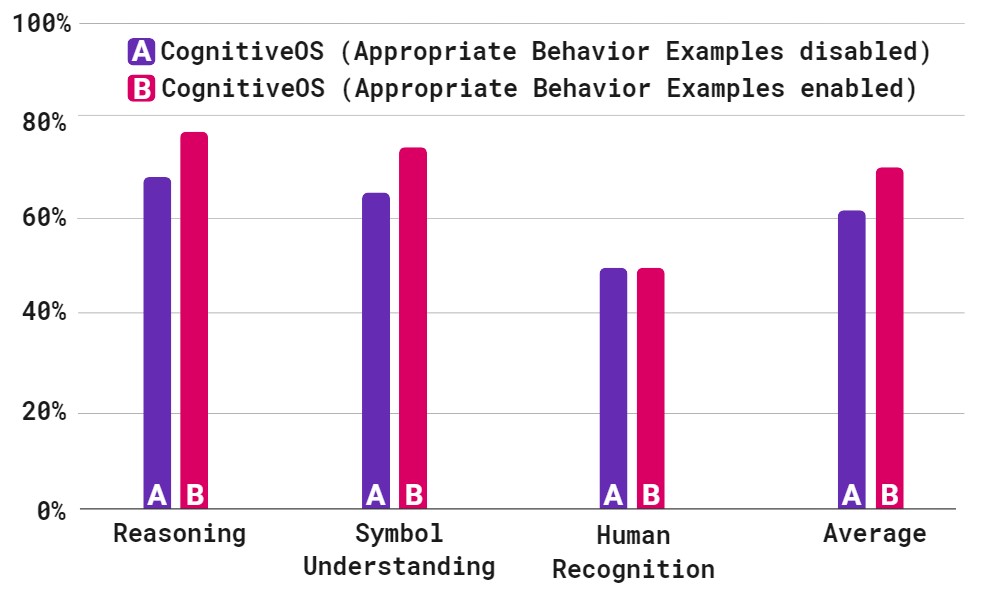}
  \caption{Performance comparison for three categories of
emergent capabilities for different types of robotic platforms.}
  \label{fig:res2}
\end{figure}

\subsection{Effect of Behavior Patterns Module}

It is well-established that one-shot learning improves the performance of LLMs and produces more structured and predictable outputs. In this experiment, we tested the system's performance on cognitively challenging tasks with and without the use of the Behavior Patterns Module. Examples of behaviors that were absent in the training dataset were included in the Behavior Patterns Vector Database. When constructing the behavior plan for the robot, the system added one-shot behavior examples for similar tasks from the database into the prompt of the Behavior Generation Module. The experimental results are illustrated in Fig. \ref{fig:res2}.

The results showed an increase in the quality of behavior generation when using the Behavior Pattern Preparation Module. The Reasoning and Symbol Understanding categories each had a 9\% benefit from using the module, while Human Recognition was not affected by it. It can be concluded that the module primarily helps the system find solutions to logically complex problems by providing the system with examples of how to reason in similar situations. In tasks aimed more at visual recognition, the module does not play a role, and to improve performance it is necessary to pay attention to the visual modules of the system. On average, the module has noticeably increased performance by 7\%.

\subsection{Robotic Collaboration Analysis}

\begin{figure}[!h]
 \centering
 \vspace{0.2cm}
  \includegraphics[width=0.97\linewidth]{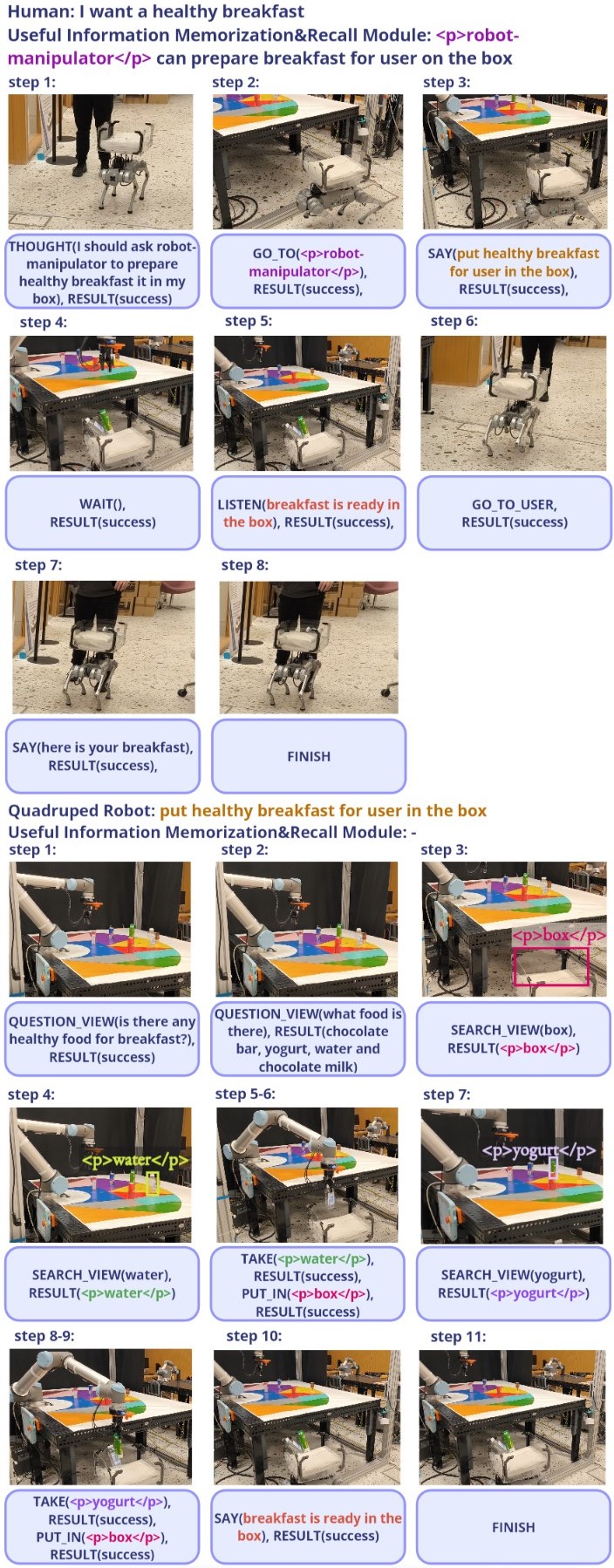}
  \vspace{-0.2cm}
  \caption{Example of cooperative interaction between two robots with CognitiveOS from the perspective of a quadruped robot and a manipulator robot.}
  \vspace{-0.4cm}
  \label{fig:res3}
\end{figure}

In this experiment with the CognitiveOS system, we provided qualitative evaluations of the extended capabilities demonstrated by a team of two robots with different morphologies through cooperation facilitated by LLM. We designed a set of tasks that supports both the standard actions, which are similar for all types of robots, and the unique actions, that can be achieved by only one of two robots in a pair, so that the whole task could be resolved only by two robots together. Each robot generates its plan independently, however, it is aware of the presence of the other robots through the Memory Module. Additionally, robots communicate with each other through the actions \textbf{“SAY”} and \textbf{“LISTEN”}. 
The example of cooperative interaction is presented in Figure \ref{fig:res3}. The tasks were conveyed by the operator to one of the two robots through voice commands. Each cognitive task required both robots to apply actions of Environmental Analysis, Object Based Question Answering, Object Localization, and Useful Information Memorization\&Recall modules. For example, the task "Pick out and give me the healthiest drink" was tested on the quadruped robot, in which case it invoked localization and speech actions, delivering this information to the manipulator. The manipulator performed object-based question answering and object localization to pick the can with orange juice and to put it on the quadruped robot's basket for delivery, vocalizing the end of operation. The successful performance of 10 such scenarios demonstrated the emerging reasoning capabilities required from the LLM systems on board of the robots to delegate tasks between other agents equipped with the required tools. Moreover, the successful performance of this task by CognitiveOS demonstrates not only its conceptual understanding of the robot agents and their potential actions, but also the successful exchange of this information by robots through conventional verbal methods of data transferring. 

\subsection{Ethical Rules Impact Assessment}

The ethical question is of paramount importance for robotics as we move towards integrating robots as universal assistants into our daily lives. In this experiment, we evaluated the impact of the Appropriate Ethical Instructions Preparing Module on the generation of robot behavior requiring adherence to specific ethical rules. We prepared a set of such tasks and assigned them to the robot with the Appropriate Ethical Instructions Preparing Module. In the former case, the system was provided with ten rules to be followed for each specific task from the Ethical Laws Vector Database and incorporated them into the prompt of the Behavior Generation Module. To evaluate the behavior of the module, we generated 200 requests, 100 that did not break the ethical rules, and 100 that broke the ethical rules, 10 for each law introduced in the dataset. The results are presented in a confusion matrix in Table \ref{tab:confusion}. Results show that 95\% of the requests that violated the ethical rules were not followed because the proposed module warned of an ethical problem, and 3\% of rules that did not violate the ethical rules were not performed because the system understood that some of the rules could be broken. These results show the high potential of the module. 

\begin{table}[h]
\centering{
\vspace{0.3cm}
\caption{Confusion Matrix for Actual and Predicted Behavior in the Ethical Analysis}
\label{tab:confusion}
\setlength{\tabcolsep}{8pt} 
\renewcommand{\arraystretch}{1.5} 
\begin{tabular}{|c c | c  c |l} \cline{1-4}
\multicolumn{2}{|c|}{}                                             & \multicolumn{2}{c|}{\textbf{Predicted Behavior}}
&  \\ \cline{3-4}
\multicolumn{2}{|c|}{\multirow{-2}{*}{}}                           & \multicolumn{1}{c|}{\textit{Negative}}                  & \textit{Positive}                                          &  \\ \cline{1-4}
\multicolumn{1}{|c|}{}                                  & \textit{Negative} & \multicolumn{1}{c|}{\cellcolor[HTML]{305496}{\color[HTML]{FFFFFF} 97}} & 3                                                 &  \\ \cline{2-4}
\multicolumn{1}{|c|}{\multirow{-2}{*}{\textbf{Actual Behavior}}} & \textit{Positive} & \multicolumn{1}{c|}{\cellcolor[HTML]{FBFCFD}5}                         & \cellcolor[HTML]{355899}{\color[HTML]{FFFFFF} 95} &  \\ \cline{1-4}
\end{tabular}}
\vspace{-0.2cm}
\end{table}



\section{Conclusion}

In this study, we introduced the first universal technology in the field of cognitive robotics. The presented CognitiveOS empowers robots of various types with cognitive capabilities through Generative AI. A key feature of the proposed development is its modular architecture, which offers a unique opportunity not only to deploy it across different robotic platforms but also to select and configure various AI-based modules without the need for a complete system overhaul. Within the scope of this research, we implemented and presented nine different modules and conducted benchmarking of the system on two platforms: a quadruped robot and a robot manipulator. The evaluation demonstrated that, on average, the system outperforms existing solutions on both platforms (for the quadruped robot and manipulator by 15\% and 10\%, respectively), showcasing exceptional results in Reasoning emergent capability, reaching previously unattainable levels of 77\%, which is 15\% higher than the second-best performance.

Evaluation of individual modules of the system revealed a significant positive impact of the behavioral example based on the robot's cognitive abilities and excellent results from the Ethical Module in correcting ethically incorrect behaviors. This overall demonstrates the viability of a modular approach to robot intelligence, where modules can be added and removed from the system as needed and for resource optimization without affecting the functionality of the rest of the system. Unlike classical end-to-end solutions, this system is highly adaptable and scalable, which allows adding modules beyond those presented. CognitiveOS can serve researchers in the field of cognitive robotics as a fundamental solution for integrating and testing their modules into it without having to build a cognitive robot from scratch.

\bibliographystyle{ieeetr}
\bibliography{root}

\end{document}